\documentclass[conference]{IEEEtran}

\usepackage{algorithmic}
\usepackage{amsmath,amssymb,amsfonts}
\usepackage{balance}
\usepackage{bm}
\usepackage{cite}
\usepackage{colortbl}
\usepackage[T1]{fontenc}
\usepackage{graphicx}
\usepackage{makecell}
\usepackage{mathtools}
\usepackage{multirow}
\usepackage{siunitx}
\usepackage{textcomp}
\usepackage{threeparttable}
\usepackage[color=green!40]{todonotes}
\usepackage{xcolor}
\usepackage{hyperref}
\graphicspath{{./}{figures/}}

\begin{document}

\title{%
  Hybrid Building/Floor Classification and Location Coordinates Regression Using
  A Single-Input and Multi-Output Deep Neural Network for Large-Scale Indoor
  Localization Based on Wi-Fi Fingerprinting%
}

\author{%
  \IEEEauthorblockN{Kyeong Soo Kim}%
  \IEEEauthorblockA{%
    Department of Electrical and Electronic Engineering\\%
    Xi'an Jiaotong-Liverpool University\\%
    Suzhou, 215123, P. R. China.\\%
    Email: Kyeongsoo.Kim@xjtlu.edu.cn%
  }%
}

\maketitle

\begin{abstract}
  In this paper, we propose hybrid building/floor classification and floor-level
  two-dimensional location coordinates regression using a single-input and
  multi-output (SIMO) deep neural network (DNN) for large-scale indoor
  localization based on Wi-Fi fingerprinting. The proposed scheme exploits the
  different nature of the estimation of building/floor and floor-level location
  coordinates and uses a different estimation framework for each task with a
  dedicated output and hidden layers enabled by SIMO DNN architecture. We carry
  out preliminary evaluation of the performance of the hybrid floor
  classification and floor-level two-dimensional location coordinates regression
  using new Wi-Fi crowdsourced fingerprinting datasets provided by Tampere
  University of Technology (TUT), Finland, covering a single building with five
  floors. Experimental results demonstrate that the proposed SIMO-DNN-based
  hybrid classification/regression scheme outperforms existing schemes in terms
  of both floor detection rate and mean positioning errors.
\end{abstract}


\section{Introduction}
\label{sec:introduction}
Of many localization techniques available nowadays, the \textit{location
  fingerprinting} is one of the most popular and promising technologies for
indoor localization \cite{he16:_wi_fi}. Because the location fingerprinting
technique does not rely on the access to line-of-sight signal from global
navigation satellite systems (GNSSs) and can be implemented based on existing
wireless infrastructure (e.g., Wi-Fi APs), it can be readily deployed without
installation of new infrastructure or modification of existing one, which is its
clear advantage over alternative techniques like \textit{triangulation} based on
time of arrival (TOA) requiring precise synchronization among all transmitters
and receivers in the system and non-standard timestamp labeling for the
measurement of distances between a target and reference points
\cite{liu07:_survey}.

In case of Wi-Fi fingerprinting, a vector of pairs of a medium access control
(MAC) address and a received signal strength (RSS) from a Wi-Fi access point
(AP) measured at a location form its \textit{location fingerprint};
the location of a user/device then can be estimated by finding the closest match
between its RSS measurement and the fingerprints of known locations in a
database \cite{bahl00:_radar}. One of the major challenges in Wi-Fi
fingerprinting is how to deal with the random fluctuation of a signal, the noise
from multi-path effects, and the device and position dependency in RSS
measurements. Recently the popular deep neural networks (DNNs) have been used in
Wi-Fi fingerprinting as well
\cite{zhang16:_deep,felix16,nowicki17:_low_wifi,Kim:18-2,Kim:18-1,adege18:_apply_dnn},
which can provide attractive solutions due to their less parameter tuning and
adaptability to a wider range of conditions with standard architectures and
training algorithms. Especially, a single-DNN-based indoor localization system
can provide a unique advantage over indoor localization systems based on
traditional machine learning techniques that, once trained, it does not need the
fingerprint database any longer but carries the necessary information for
localization in DNN weights and biases, which could enable a secure and
energy-efficient indoor localization exclusively running on mobile devices
without exchanging any data with the fingerprint server \cite{Kim:18-1}.

When we need to estimate a location in a large \textit{building complex} like a
big shopping mall or a university campus, the scalability of fingerprinting
schemes becomes a major issue, too. The current state-of-the-art Wi-Fi
fingerprinting techniques adopt a hierarchical approach, where the building,
floor, and position (e.g., a label or coordinates) of a location are estimated
in a hierarchical and sequential manner using \textit{possibly} a different
algorithm tailored for each task \cite{moreira15:_wi_fi}. The application of
this hierarchical and sequential approach for multi-building and multi-floor
indoor localization to DNN-based schemes, however, may cause scalability issues:
As discussed in \cite{Kim:18-1}, compared to the traditional techniques as
proposed in \cite{moreira15:_wi_fi}, DNNs for different levels of localization
need to be trained separately with either a system-wide dataset (i.e., a DNN for
building estimation) or multiple sub-datasets derived from the common dataset
(i.e., building-specific datasets for DNNs for floor estimation and
building-floor-specific datasets for DNNs for building-floor-level location
estimation), which poses significant challenges on the management of location
fingerprint databases as well as the training of possibly a large number of
DNNs.

\begin{figure}[!tbp]
  \begin{center}
    \includegraphics[angle=-90,width=\linewidth]{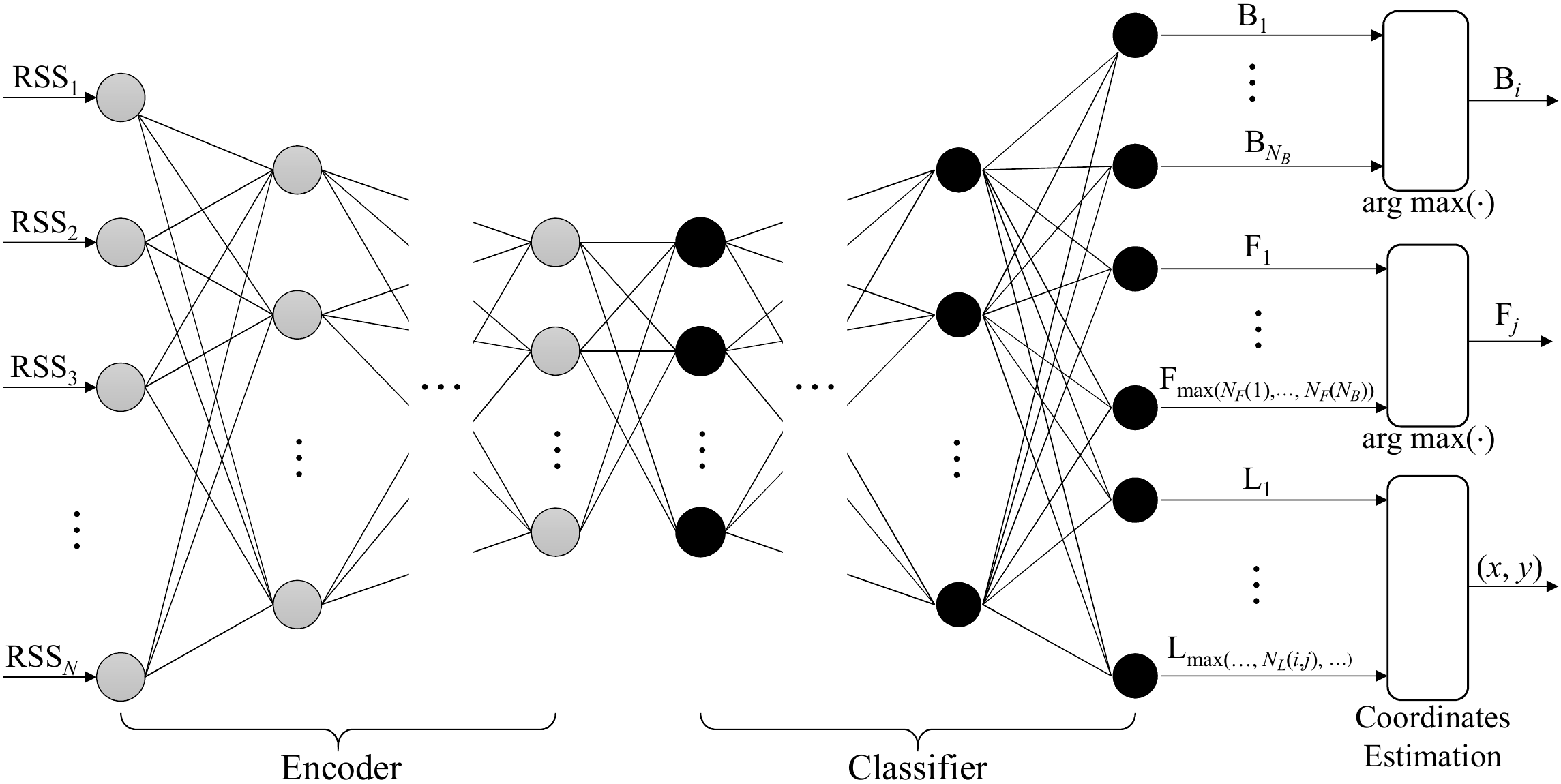}
  \end{center}
  \caption{A DNN architecture for scalable multi-building and multi-floor indoor
    localization based on an stacked autoencoder (SAE) for the reduction of
    feature space dimension and a feed-forward classifier for multi-label
    classification \cite{Kim:18-1}.}
  \label{fig:scalable_dnn_classifier}
\end{figure}

To address the scalability issue of DNN-based multi-building and multi-floor
indoor localization, the scalable DNN architecture based on \textit{multi-label
  classification} \cite{tsoumakas07:_multi} shown in
Figure~\ref{fig:scalable_dnn_classifier} was proposed in \cite{Kim:18-1}, which
can greatly reduce the number of output nodes compared to that of the DNN
architecture based on multi-class classification.
This DNN architecture also enables customized processing of parts of DNN outputs
for building, floor, and location (i.e., the functional blocks on the right in
Figure~\ref{fig:scalable_dnn_classifier}) due to the straightforward mapping
between building, floor, and location identifiers and its corresponding
one-hot-encoded categorical variable.
%

In this paper, to further exploit the hierarchical nature of multi-building and
multi-floor indoor localization, we study the extension of the scalable DNN
architecture proposed in \cite{Kim:18-1} based on single-input and multi-output
(SIMO) DNN architecture, a special case of more general multi-input and
multi-output (MIMO) DNN architecture \cite{keras-mimo}; this SIMO-DNN-based
extension enables hybrid building/floor classification and floor-level
two-dimensional location coordinates regression through a dedicated output for
each task, which can take into account the different nature of the estimation of
building/floor and floor-level coordinates.

The rest of the paper is organized as follows: In
Sec.~\ref{sec:location-estimation}, we revisit the problem of location
coordinates estimation in multi-building and multi-floor indoor localization and
consider the two options of classification and regression. In
Sec.~\ref{sec:simo-dnn}, we propose a new multi-building and multi-floor indoor
localization scheme based on SIMO-DNN-based hybrid
classification/regression. Sec.~\ref{sec:experimental-restuls} presents
experimental results for the localization performance of the proposed
SIMO-DNN-based hybrid classification/regression
scheme. Sec.~\ref{sec:conclusions} concludes our work in this paper.

\section{Location Coordinates Estimation in Multi-Building and Multi-Floor
  Indoor Localization: Classification vs Regression}
\label{sec:location-estimation}
In \cite{lohan17:_wi_fi}, the authors present a new crowdsourced Wi-Fi
fingerprint database\footnote{We call it TUT database from now on in this
  paper.} comprised of 4648 fingerprints collected with 21 devices covering the
five floors of a six-floor building in Tampere University of Technology,
Finland, which is publicly available and hosted in public EU Zenodo
repository. Compared to UJIIndoorLoc database \cite{torres-sospedra14:_ujiin},
i.e., another well-known public fingerprint database covering three buildings
with four floors each, the fingerprints of TUT database were collected around a
single building, but they provide three-dimensional coordinates (i.e.,
$(x, y, z)$ of a measurement reference point); the availability of
three-dimensional coordinates is indeed a major reason that we use the TUT
database in this paper, which motivated us to investigate DNN-based regression
of location coordinates.

In \cite{Kim:18-1}, the DNN-based multi-building and multi-floor indoor
localization is done based on multi-label classification of building, floor and
labeled position (i.e., \textit{reference points} in the UJIIndoorLoc database
training subset). The two-dimensional coordinates of an unknown position is then
determined by a weighted average of the coordinates of multiple candidate
reference points through the procedure described in Fig.~9 of
\cite{Kim:18-1}. Direct regression of location coordinates with DNNs could
eliminate such additional procedure.

There is also another reason that we consider a regression-based approach for
the TUT database. If we apply classification for the estimation of location as in
\cite{Kim:18-1}, we need multiple fingerprint samples per label (i.e., reference
point) to train DNNs. This is the case for the UJIIndoorLoc database training
subset, where there are about 21 fingerprint samples per reference point in
average ($\approx$19674\,samples$\div$933\,reference\,points). Because the
fingerprints in the TUT database are not collected at fixed reference points
(e.g., office, lab, and corridor) or grid points but at any points inside and
outside the building, however, there are few fingerprint samples per measurement
point; for its training subset of 697 fingerprint samples, there are 694 unique
locations, which results in 1.004 samples per reference point in
average. Therefore, regression is the only viable option for floor-level
location coordinates estimation with the TUT database.\footnote{To apply
  classification, we regrouped multiple fingerprint samples based on grids using
  their coordinates, but its results are not as good as those based on hybrid
  classification/regression approach described in this paper.}

Regarding floor estimation, we consider two options, i.e., pure regression of
three-dimensional location coordinates and hybrid floor classification and
regression of two-dimensional location coordinates: If we treat the $z$
coordinate of a location as exactly as the $x$ and $y$ coordinates, we can apply
pure regression for three-dimensional coordinates. If we treat the $z$
coordinate as a label (i.e., multiples of 3.7 --- $0, 3.7, 7.4, 11.1, 14.8$ ---
for five floors), on the other hand, we can apply classification for floor
estimation, while $(x, y)$ coordinates estimation is still done by usual
regression. Compared to the pure regression of three-dimensional location
coordinates, we can better exploit the hierarchical nature of multi-building and
multi-floor indoor localization with the hybrid classification/regression by
separate processing of the information at different levels.

These considerations lead us to a SIMO DNN architecture described in
Sec.~\ref{sec:simo-dnn}, which enables the hybrid classification/regression
approach.

\section{SIMO DNN for Hybrid Classification and Regression}
\label{sec:simo-dnn}
\begin{figure}[!tbp]
  \begin{center}
    \includegraphics[angle=-90,width=.8\linewidth]{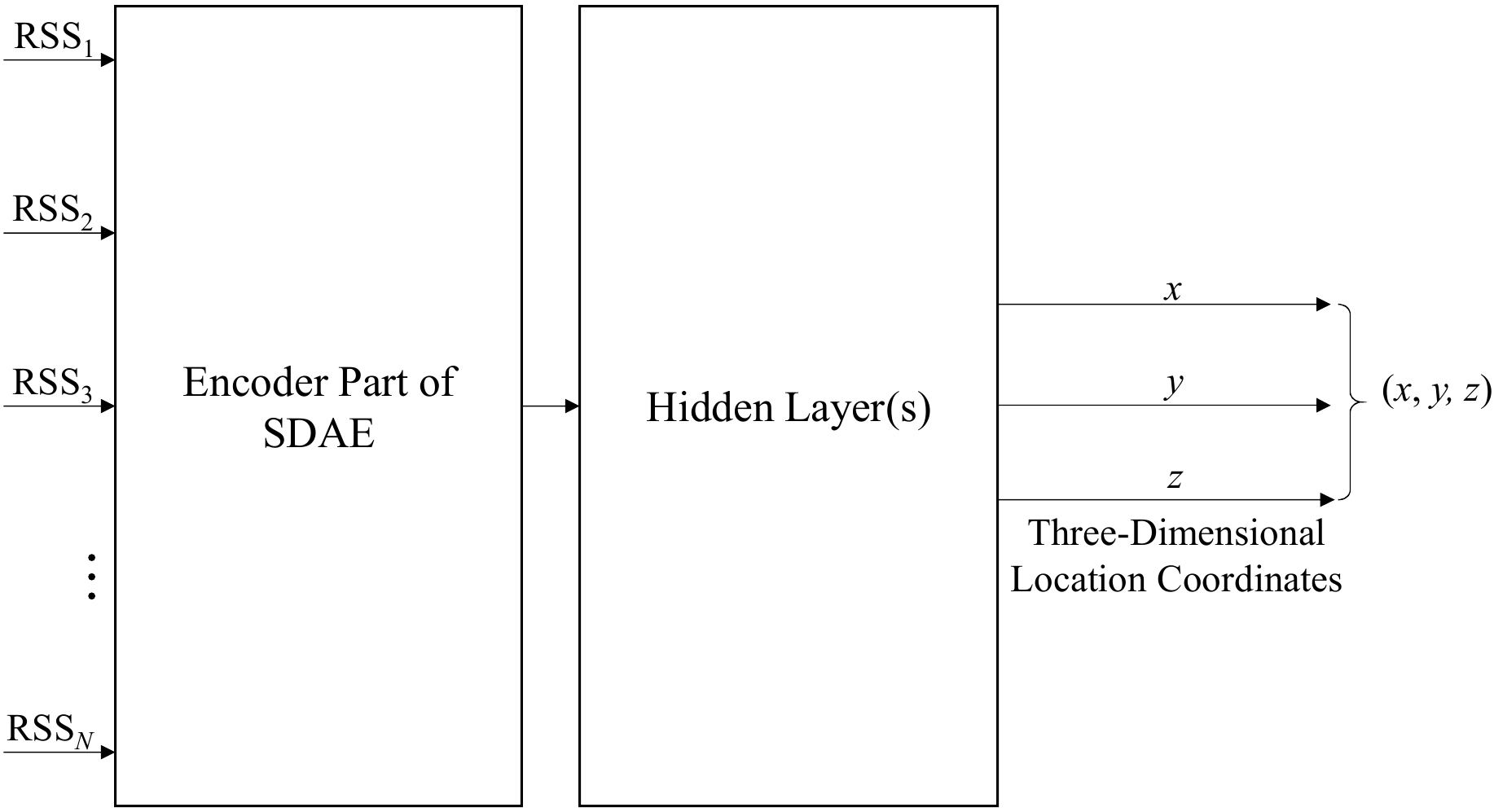}
  \end{center}
  \caption{A SISO DNN architecture for three-dimensional coordinates regression
    for multi-building and multi-floor indoor localization.}
  \label{fig:siso_dnn_regression}
\end{figure}

Given the availability of three-dimensional coordinates of reference points in
the TUT database, one can come up with an indoor localization scheme based on
the DNN-based coordinates regression shown in
Fig.~\ref{fig:siso_dnn_regression}, which serves as a reference scheme in this
paper.
This single-input and single-output (SISO\footnote{The definitions of SISO,
  SIMO, and MIMO DNN architectures are implementation-oriented rather than
  mathematical; even with SISO, we can have multiple input values (i.e., a
  vector-valued input), which, however, are grouped together with a common loss
  function and a loss weight.})-DNN-based three-dimensional coordinates
regression scheme, however, treats all three coordinates equal and thereby
cannot take into account the discrete nature of $z$ coordinate (i.e., multiples
of 3.7) and its relation to the floor estimation, which should be given priority
over the other two coordinates.



\begin{figure}[!tbp]
  \begin{center}
    \includegraphics[angle=-90,width=\linewidth,trim=0 1 0 2,clip=true]{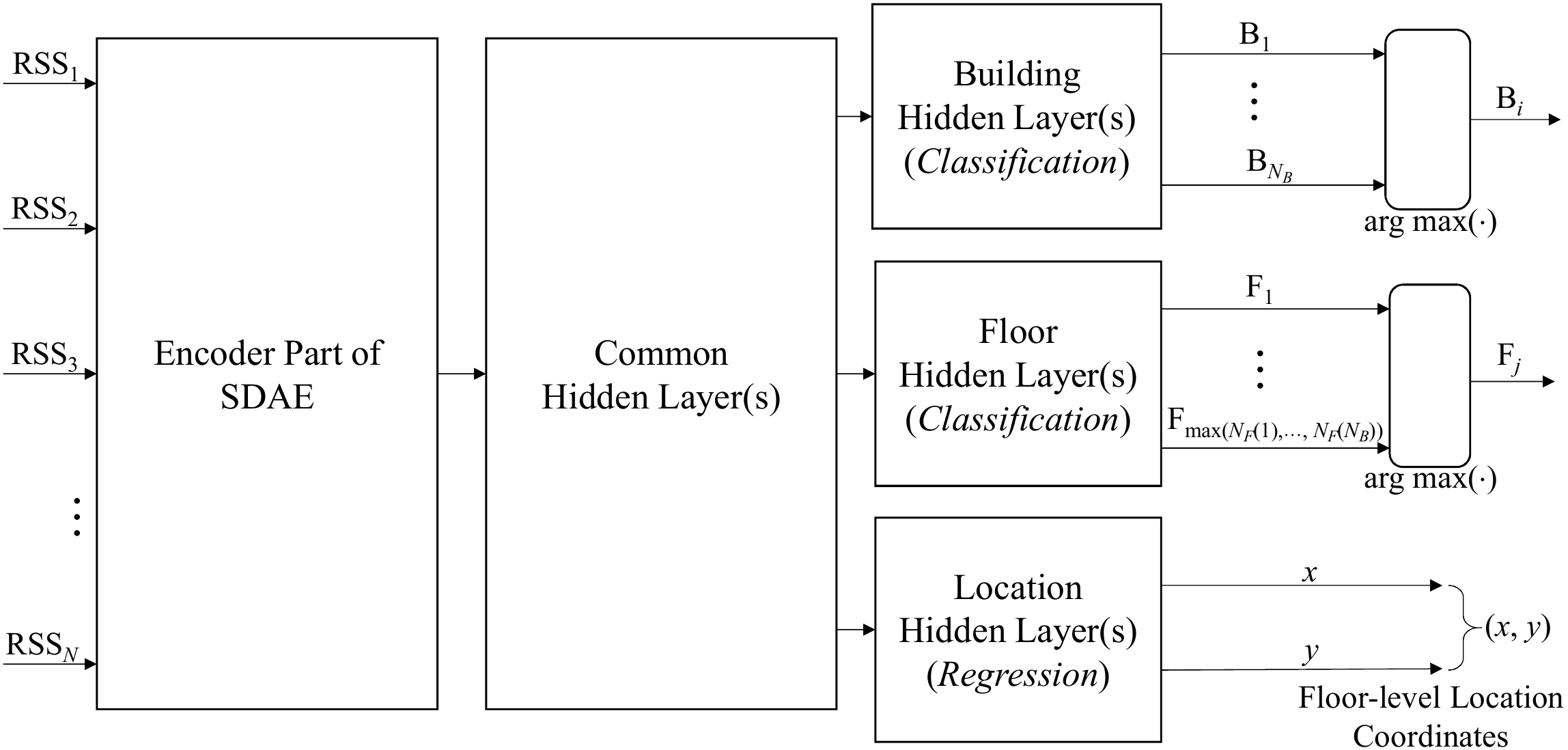}
  \end{center}
  \caption{A SIMO DNN architecture for hybrid building/floor classification and
    floor-level two-dimensional coordinates regression.}
  \label{fig:simo_dnn_hybrid}
\end{figure}

As shown in Fig.~\ref{fig:simo_dnn_hybrid}, on the other hand, the SIMO DNN
architecture enables the use of a different estimation framework for a different
sub-problem; with a separate output and hidden layers dedicated for a
sub-problem, we can use different activation and loss functions optimized for
the choice of estimation framework. For example, we can use \textit{softmax}
activation function and \textit{categorical crossentropy} loss function for
multi-class classification of floor at the floor output, while we can use
\textit{linear} activation function and \textit{mean squared error (MSE)} loss
function for regression of location coordinates at the location output.

Note that in the proposed SIMO DNN architecture, the SAE of the single-DNN
architecture proposed in \cite{Kim:18-1} (also shown in
\ref{fig:scalable_dnn_classifier}) is replaced by the stacked denoising
autoencoder (SDAE) based on the results in \cite{vincent10:_stack}, where the
authors argue that \textit{denoising autoencoder pretraining} provides better
classification performance than \textit{ordinary autoencoder pretraining}
because the denoising criterion as a tractable unsupervised objective enables
DNNs to learn more useful higher-level representations. Our own experimental
investigation also confirms this claim.

\section{Experimental Results}
\label{sec:experimental-restuls}
To evaluate the localization performance of the proposed SIMO-DNN-based hybrid
classification/regression scheme, we carry out experiments using the new TUT
Wi-Fi fingerprinting database \cite{lohan17:_wi_fi}, covering a single building
with five floors. Both SISO (reference) and SIMO (proposed) DNN models are
implemented based on Keras \cite{keras} and TensorFlow \cite{tensorflow}.

We use \textit{EarlyStopping} together with \textit{ModelCheckpoint} callbacks
of Keras to save the best weights and biases during the training phase and use
them for the performance evaluation with a test dataset. Each simulation run is
repeated twenty times with different random number seeds to calculate a 95\%
confidence interval. Tables~\ref{tbl:simo_dnn_parameters} and
\ref{tbl:siso_dnn_parameters} summarizes DNN parameter values, which are chosen
experimentally and used throughout the experiments.
\begin{table*}[!tbp]
  \centering
  \begin{threeparttable}
    \caption{SIMO DNN Parameter Values for Hybrid Classification/Regression.}
    \label{tbl:simo_dnn_parameters}
    \begin{tabular}{ll}
      \hline
      \multicolumn{1}{c}{DNN Parameter} & \multicolumn{1}{c}{Value} \\ \hline
      Fraction of training data used as validation data & 0.2 \\
      Number of Epochs\tnote{1} & 100 \\
      Batch Size & 64 \\
      Optimizer & Nesterov-accelerated Adaptive Moment Estimation (NADAM)
                  \cite{dozat16:_incor_nester_adam} \\ \Xhline{.5\arrayrulewidth}
      SDAE Hidden Layers & 1024-1024-1024 \\
      SDAE Activation & Sigmoid \\
      SDAE Corruption Level & 0.1 \\
      SDAE Loss & Mean Squared Error (MSE) \\ \Xhline{.5\arrayrulewidth}
      Common Hidden Layer & 1024 \\
      Common Hidden Layer Activation & Rectified Linear (ReLU) \\
      Common Hidden Layer Dropout Rate & 0.25 \\ \Xhline{.5\arrayrulewidth}
      Floor Classifier Hidden Layer & 256 \\
      Floor Classifier Hidden Layer Activation & ReLU \\
      Floor Classifier Hidden Layer Dropout Rate & 0.25 \\
      Floor Classifier Output Layer Activation & Softmax \\
      Floor Classifier Loss & Categorical Crossentropy \\ \Xhline{.5\arrayrulewidth}
      Coordinates Regressor Hidden Layer & 256 \\
      Coordinates Regressor Hidden Layer Activation & ReLU \\
      Coordinates Regressor Hidden Layer Dropout Rate & 0.25 \\
      Coordinates Regressor Output Layer Activation & Linear \\
      Coordinates Regressor Loss & MSE \\ \hline
    \end{tabular}
    \begin{tablenotes}
    \item[1] With Keras \textit{EarlyStopping} (min\_delta=0 and patience=10)
      and \textit{ModelCheckpoint} callbacks.
    \end{tablenotes}
  \end{threeparttable}
\end{table*}
\begin{table*}[!tbp]
  \centering
  \begin{threeparttable}
    \caption{SISO DNN Parameter Values for Three-Dimensional Coordinates
      Regression.}
    \label{tbl:siso_dnn_parameters}
    \begin{tabular}{ll}
      \hline
      \multicolumn{1}{c}{DNN Parameter} & \multicolumn{1}{c}{Value} \\ \hline
      Fraction of training data used as validation data & 0.2 \\
      Number of Epochs\tnote{1} & 100 \\
      Batch Size & 64 \\
      Optimizer & NADAM \cite{dozat16:_incor_nester_adam} \\ \Xhline{.5\arrayrulewidth}
      SDAE Hidden Layers & 1024-1024-1024 \\
      SDAE Activation & Sigmoid \\
      SDAE Corruption Level & 0.1 \\
      SDAE Loss & MSE \\ \Xhline{.5\arrayrulewidth}
      Hidden Layer & 1024 \\
      Hidden Layer Activation & ReLU \\
      Hidden Layer Dropout Rate & 0.25 \\
      Output Layer Activation & Linear \\
      Coordinates Regressor Loss & MSE \\ \hline
    \end{tabular}
    \begin{tablenotes}
    \item[1] With Keras \textit{EarlyStopping} (min\_delta=0 and patience=10)
      and \textit{ModelCheckpoint} callbacks.
    \end{tablenotes}
  \end{threeparttable}
\end{table*}

\begin{figure*}[!tbp]
  \newlength{\mywidth}%
  \setlength{\mywidth}{.32\textwidth}%
  \begin{minipage}[c]{\mywidth}
    \begin{center}
      \includegraphics[width=\linewidth,trim=10 8 45 30,clip=true]{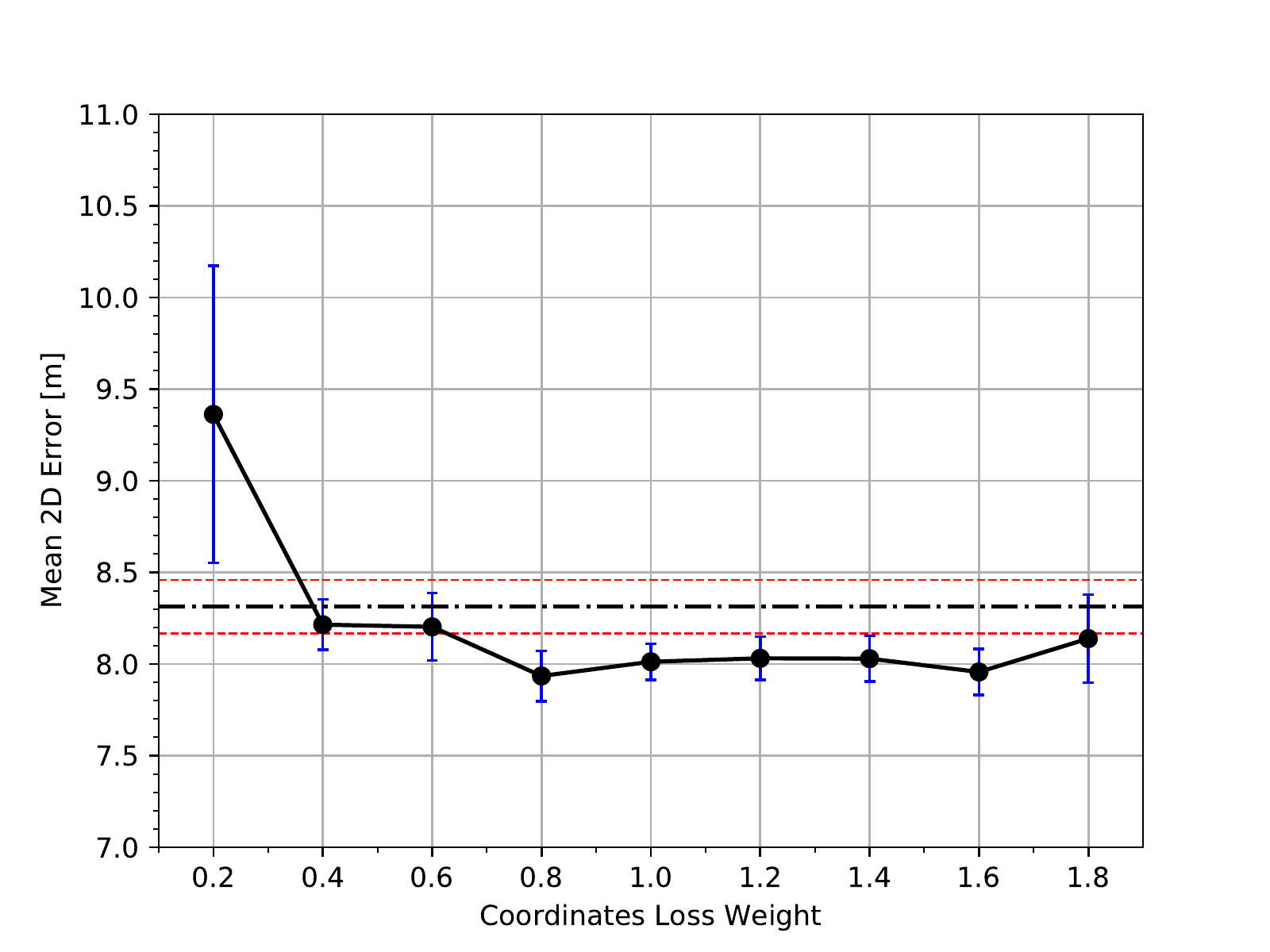}\\
      {\scriptsize (a)}
    \end{center}
  \end{minipage}
  \hfill
  \begin{minipage}[c]{\mywidth}
    \begin{center}
      \includegraphics[width=\linewidth,trim=10 8 45 30,clip=true]{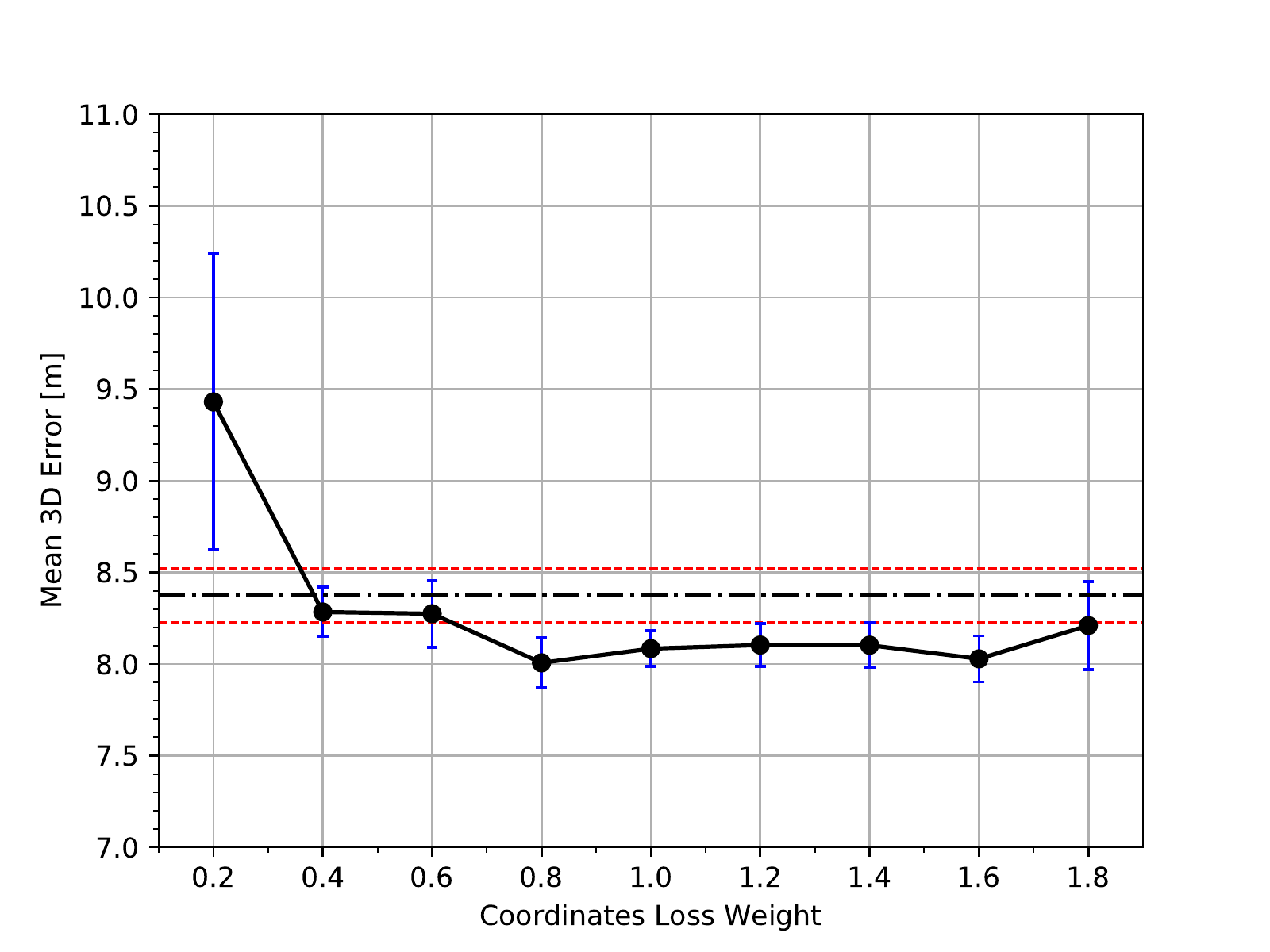}\\
      {\scriptsize (b)}
    \end{center}
  \end{minipage}
  \hfill
  \begin{minipage}[c]{\mywidth}
    \begin{center}
      \includegraphics[width=\linewidth,trim=10 8 45 30,clip=true]{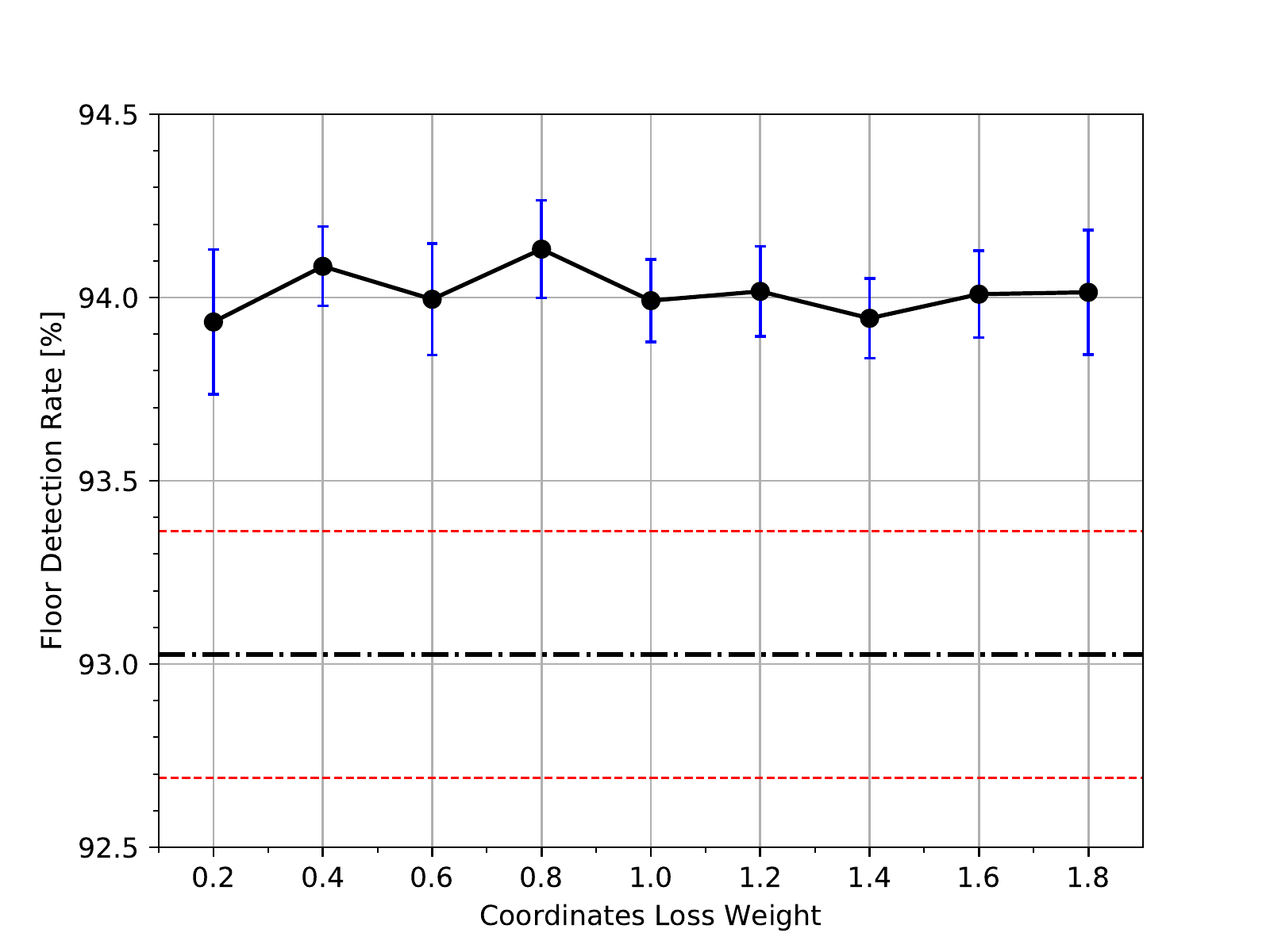}\\
      {\scriptsize (c)}
    \end{center}
  \end{minipage}
  \caption{Effect of coordinates loss weight (with fixed floor loss weight of
    1.0) on the localization performance of SIMO-DNN-based hybrid
    classification/regression: (a) Mean two-dimensional positioning error [m],
    (b) mean three-dimensional positioning error [m], and (c) floor detection
    rate [\%].}
  \label{fig:clw_effect}
\end{figure*}

Fig.~\ref{fig:clw_effect} shows the effect of coordinates loss weight on the
localization performance of the proposed SIMO-DNN-based hybrid
classification/regression scheme, where we plot mean two-dimensional positions
error, mean three-dimensional positioning error, and floor detection rate, all
with 95\% confidence intervals. For a comparison, we also show the localization
performance of the SISO-based three-dimensional coordinates regression as three
horizontal lines (i.e., the dash-dotted line in the middle for a mean value and
the two dash lines for a 95\% confidence interval).

The mean two-dimensional and three-dimensional positioning errors shown in
Fig.~\ref{fig:clw_effect}~(a) and (b) indicate that the proposed SIMO-DNN-based
hybrid classification/regression scheme outperforms the reference SISO-DNN-based
regression scheme for a wide range of coordinates loss weight; in case of floor
detection rate shown in Fig.~\ref{fig:clw_effect}~(c), the proposed scheme
provides better performance than the reference one (i.e., around 1\% higher
floor detection rate) irrespective of the coordinates loss weight. The detailed
investigation of the results shows that coordinates loss weight of 0.8 (with
floor loss weight of 1.0) provides the best overall performance. Also, from the
overall results shown in Fig.~\ref{fig:clw_effect}, we observe that the use of
proper estimation framework for a given task enabled by the SIMO DNN
architecture is more important than the loss weight control of multiple outputs.

Table~\ref{tab:comparison_tut_best} compares the localization performance of the
proposed and reference schemes with the best results from the benchmark
positioning results in \cite{lohan17:_wi_fi}.
\begin{table*}[!tb]
  \begin{threeparttable}
    \definecolor{light-gray}{gray}{0.75}
    \caption{Comparison of the localization performance of the proposed scheme
      with that of SISO-DNN-based three-dimensional regression and the best
      results from the benchmark in \cite{lohan17:_wi_fi}.}
    \label{tab:comparison_tut_best}
    \centering
    \begin{tabular}{lcccc}
      \hline
      \multicolumn{1}{c}{Algorithm} & Mean 2D Error [m] & Mean 3D Error [m] & Floor Detection [\%] & Notes \\ \hline
      \textbf{SIMO-DNN-Based Hybrid Classification/Regression}\tnote{1} & \textbf{7.46}\tnote{2} & \textbf{7.53}\tnote{2} & \textbf{94.53}\tnote{3} & Proposed scheme. \\
      SISO-DNN-Based 3D Regression & 7.88\tnote{2} & 7.94\tnote{2} & 94.20\tnote{3} & Reference scheme. \\
      RSS Clustering (Affinity Propagation) \cite{cramariuc16:_clust_wifi} & 8.09 & 8.70 & 90.81 & From \cite{lohan17:_wi_fi}. \\
      UJI kNN Algorithm\tnote{4} \cite{torres-sospedra15:_compr_wi_fi} & 8.65 & 8.92 & 92.99 & From \cite{lohan17:_wi_fi}. \\
                                    & & & \\ \hline
    \end{tabular}
    \begin{tablenotes}
    \item[1] With floor loss weight=1.0 and coordinates loss weight=0.8.
    \item[2] Minimum value from 20 runs.
    \item[3] Maximum value from 20 runs.
    \item[4] With data=powed, dist=sorensen, N\textsubscript{nn}=1,
      Not\textsubscript{heard}=-103 \cite{lohan17:_wi_fi}.
    \end{tablenotes}
  \end{threeparttable}
\end{table*}
Even though the TUT dataset division is more challenging than other available
Wi-Fi datasets, by having only 15\% of samples for training/reference, compared
to 85\% of samples for evaluation, the proposed SIMO-DNN-based hybrid
classification/regression scheme outperforms the best algorithms from the
benchmark in \cite{lohan17:_wi_fi} in all three categories, which is remarkable
considering that DNN-based schemes require lots of training data compared to
traditional machine learning techniques.

Note that the results presented in this section are preliminary and only with
the TUT database; the current work is focused on the feasibility of the proposed
SIMO-DNN-based hybrid classification/regression scheme in comparison with the
SISO-DNN-based pure regression scheme and the state-of-the-art Wi-Fi
fingerprinting techniques.

\section{Conclusions}
\label{sec:conclusions}
In this paper we have proposed SIMO-DNN-based hybrid building/floor
classification and floor-level two-dimensional location coordinates regression
for large-scale indoor localization based on Wi-Fi fingerprinting. This hybrid
approach for indoor localization enabled by SIMO DNN architecture can better
exploit the hierarchical and different nature of the estimation of
building/floor and floor-level location coordinates.

The experimental results with the TUT database demonstrate the advantages of the
proposed scheme, which can provide the best overall performance in terms of mean
two-dimensional and three-dimensional positioning errors and floor detection
rate in comparison to the best algorithms from the benchmark in
\cite{lohan17:_wi_fi} as well as the reference scheme based on SISO-DNN-based
three-dimensional coordinates regression.

The results presented in this paper suggest that the proper use of estimation
frameworks tailored for given sub-problems (i.e., multi-class classification for
building/floor estimation and regression for floor-level two-dimensional
coordinates estimation) enabled by SIMO DNN architecture can address the
challenging aspects of the TUT database, including just one sample per reference
point (compared to tens or hundreds in other databases) and the unusual split
ratio between training/reference and evaluation samples (i.e., 15:75).

\section*{Acknowledgment}
This work was supported in part by Xi'an Jiaotong-Liverpool University (XJTLU)
Research Development Fund (under Grant RDF-16-02-39), Research Institute for
Future Cities Research Leap Grant Programme 2016-2017 (under Grant RIFC2018-3),
and Centre for Smart Grid and Information Convergence.

\balance 

\bibliographystyle{IEEEtran}
\bibliography{IEEEabrv,kks}

\end{document}